\documentclass[sigconf]{acmart}

\renewcommand\footnotetextcopyrightpermission[1]{}
\setcopyright{none}
\acmDOI{}
\acmISBN{}
\acmYear{}
\copyrightyear{}
\acmConference{}{}{}
\acmBooktitle{}

\usepackage{xcolor}

\newif\ifshowchanges
\showchangestrue  

\newcommand{\add}[1]{\ifshowchanges{\color{black}#1}\else#1\fi}

\usepackage{amsmath,amsfonts,bm}

\def\eqref#1{equation~\ref{#1}}

\def\1{\bm{1}}

\ifdefined\mathsfit
  \let\mathsfit\relax
\fi
\DeclareMathAlphabet{\mathsfit}{\encodingdefault}{\sfdefault}{m}{sl}
\SetMathAlphabet{\mathsfit}{bold}{\encodingdefault}{\sfdefault}{bx}{n}

\usepackage{booktabs}
\usepackage{dblfloatfix} 
\usepackage{amsfonts}
\usepackage{nicefrac}
\usepackage{microtype}
\usepackage{xcolor}
\usepackage{float}
\usepackage{tabularx}

\usepackage{colortbl}
\usepackage{array}
\usepackage[table]{xcolor}

\usepackage{threeparttable}
\usepackage{graphicx}  
\usepackage{multirow}
\usepackage{subcaption}

\usepackage{longtable}
\usepackage{algorithm}
\usepackage[noend]{algpseudocode}
\usepackage{amsmath}

\floatname{algorithm}{Algorithm}

\usepackage{listings}
\usepackage{xcolor}

\usepackage[most]{tcolorbox}
\tcbuselibrary{listings, breakable} 

\newtcblisting{mycode}[2]{
    breakable,
    arc=2pt,
    boxrule=0.5pt,
    colback=gray!5,
    colframe=gray!50,
    top=0pt,                          
    bottom=0pt,                       
    middle=0pt,                       
    listing only, 
    listing options={
        language=#1,
        basicstyle=\ttfamily\footnotesize, 
        lineskip=-1pt,                  
        breaklines=true,
        showstringspaces=false
    },
    title=#2
}

\title{Understand Then Memory: A Cognitive Gist-Driven RAG Framework with Global Semantic Diffusion}

\author{Pengcheng Zhou}
\authornote{Both authors contributed equally to this research.}
\affiliation{
  \institution{National University of Singapore}
  \country{Singapore}
}
\email{zhoupengcheng0219@gmail.com}

\author{Haochen Li}
\authornotemark[1]
\affiliation{
  \institution{Tsinghua University}
  \country{China}
}
\email{hc.li21@outlook.com}

\author{Zhiqiang Nie}
\authornotemark[1]
\affiliation{
  \institution{Tsinghua University}
  \country{China}
}
\email{nzq24@mails.tsinghua.edu.cn}

\author{JiaLe Chen}
\affiliation{
  \institution{Tsinghua University}
  \country{China}
}
\email{18838925520@163.com}

\author{Qing Gong}
\affiliation{
  \institution{Nanyang Technological University}
  \country{Singapore}
}
\email{gongqing0627@gmail.com}

\author{Weizhen Zhang}
\affiliation{
  \institution{Beijing University of Posts and Telecommunications}
  \country{China}
}
\email{zhangweizhende1256@163.com}

\author{Chun Yu}
\authornote{Corresponding author.}
\affiliation{
  \institution{Tsinghua University}
  \country{China}
}
\email{chunyu@tsinghua.edu.cn}

\begin{document}

\begin{abstract}
Retrieval-Augmented Generation (RAG) effectively mitigates hallucinations in LLMs by incorporating external knowledge. However, the inherent discrete representation of text in existing frameworks often results in a loss of semantic integrity, leading to retrieval deviations. Inspired by the human episodic memory mechanism, we propose CogitoRAG, a RAG framework that simulates human cognitive memory processes. The core of this framework lies in the extraction and evolution of the Semantic Gist. During the offline indexing stage, CogitoRAG first deduces unstructured corpora into gist memory corpora, which are then transformed into a multi-dimensional knowledge graph integrating entities, relational facts, and memory nodes. In the online retrieval stage, the framework handles complex queries via Query Decomposition Module that breaks them into comprehensive sub-queries, mimicking the cognitive decomposition humans employ for complex information. Subsequently, Entity Diffusion Module performs associative retrieval across the graph, guided by structural relevance and an entity-frequency reward mechanism. Furthermore, we propose the CogniRank algorithm, which precisely reranks candidate passages by fusing diffusion-derived scores with semantic similarity. The final evidence is delivered to the generator in a passage-memory pairing format, providing high-density information support. Experimental results across five mainstream QA benchmarks and multi-task generation on GraphBench  demonstrate that CogitoRAG significantly outperforms state-of-the-art RAG methods, showcasing superior capabilities in complex knowledge integration and reasoning.
\end{abstract}

\maketitle

\section{Introduction}

Retrieval-Augmented Generation (RAG), as a key technique for connecting external knowledge with Large Language Models (LLMs), has become the dominant paradigm by dynamically retrieving relevant information to alleviate model hallucinations~\citep{lewis2020retrieval,guu2020retrieval}. A typical RAG system generally follows a three-step pipeline: knowledge construction, retrieval, and reranking~\citep{gao2023retrieval}.

\begin{figure}[t]
  \centering
  \vspace{0.5cm}
  \includegraphics[width=\linewidth]{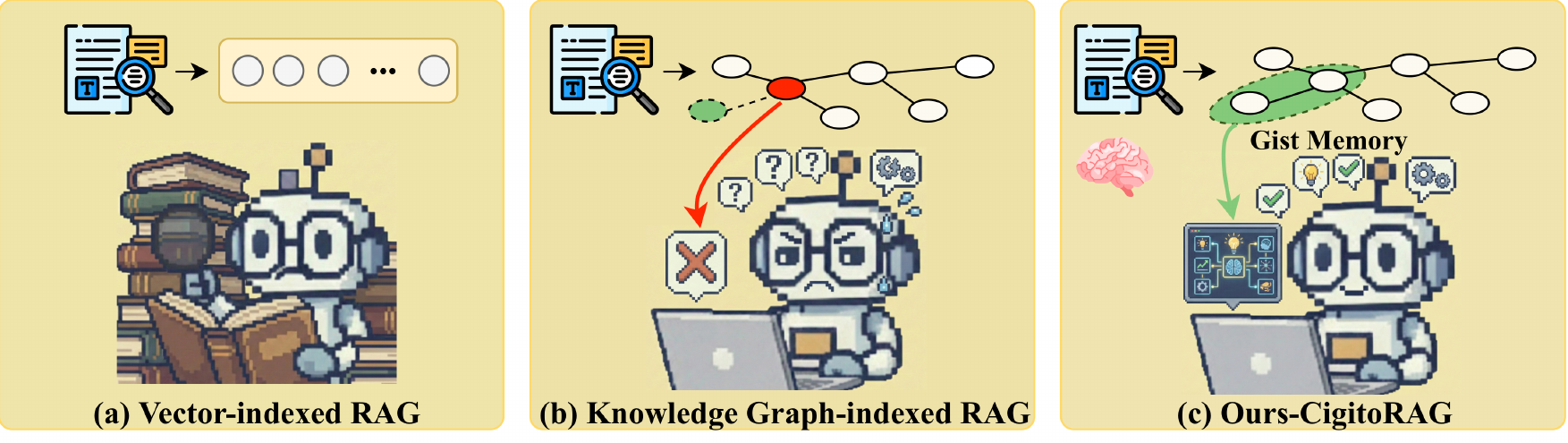}
  \caption{(a) Vector-indexed RAG: relies on local semantic matching of document chunks. (b) Knowledge Graph-indexed RAG: captures explicit entity links, limited to local triple reasoning. (c) Our CogitoRAG: extracts gist memory to capture explicit facts and implicit logic, enabling holistic understanding.}
  \label{fig:MDKG}
  \vspace{-0.5cm}
\end{figure}

To structure the knowledge base in the construction phase, recent efforts have shifted from simple text indexing~\citep{lewis2020retrieval,pmlr-v119-guu20a,pmlr-v162-borgeaud22a} to building structured knowledge graphs~\citep{saleh2024sg,huang2025ket,jiang2025hykge,guo2024lightrag}. However, this initial stage is, in essence, a lossy compression process~\citep{li2024simple,zhu2025knowledge}, which discards substantial context and narrative background, leading to a deficit in comprehension and memory. During the retrieval stage, these deficiencies are further magnified. Even with the introduction of advanced iterative~\citep{wangcomplex} or agentic~\citep{maragheh2025arag,ravuru2024agentic} reasoning frameworks, these systems can comprehend the associations between entities, they fail to understand how these associations collectively constitute a meaningful semantic scene, thereby becoming trapped in localized reasoning~\citep{fountas2024human,gutierrez2025from}. While advanced reranking techniques exist~\citep{sun2025dynamicrag,yang2025reranking,yu2024rankrag,jin2025sara}, they remain largely focused on evaluating the local relationship between a query and its candidate documents, failing to fundamentally resolve the limitations originating from flawed knowledge base construction and localized retrieval, as illustrated in Figure \ref{fig:MDKG}.

In contrast to the localized reasoning of current systems, the human brain seamlessly integrates vast amounts of knowledge. This remarkable ability is explained by well-established cognitive theories of memory. Foundational research distinguishes memory for verbatim details from ``gist memory'', a more durable understanding of semantic essence~\citep{reyna1995fuzzy}. Furthermore, ``episodic memory'' allows for the holistic comprehension of information by situating it within a coherent spatiotemporal scene~\citep{tulving1972episodic}. During recall, the brain then employs ``importance judgment'' to rapidly identify core concepts from this landscape~\citep{kintsch1978toward}.

Inspired by this, we propose CogitoRAG, a novel RAG framework that simulates the process of human cognitive memory. To emulate the core characteristics of human gist memory, CogitoRAG first infers and comprehends unstructured corpora to derive a gist memory corpus, which is then encoded and transformed into a multi-dimensional knowledge graph integrating entities, relational facts, and memory nodes, thereby fully preserving the rich contextual associations and semantic logic among knowledge units. For complex queries in the online retrieval phase, the framework decomposes them into comprehensive sub-queries via the Query Decomposition Module, accurately simulating the cognitive decomposition mechanism employed by humans when processing complex information. Building on this, we design the Entity Diffusion Module, a unified mechanism that simultaneously emulates the scene integration capability of human episodic memory and the core logic of importance judgment. By quantitatively evaluating the topological relevance and entity frequency distribution of each node, this module realizes the diffusion and propagation of semantic importance within the knowledge graph. To achieve precise screening of candidate information, we further propose the CogniRank algorithm, which re-ranks memory fragments and passages by integrating semantic space relevance and graph topological importance through a weighted fusion strategy. Ultimately, high-density evidential support is delivered to the generator in the form of passage-memory pairs. Experimental results demonstrate that on five mainstream QA benchmarks covering both single-hop and multi-hop question-answering tasks, as well as on the GraphBench multi-task generation tasks, CogitoRAG significantly outperforms existing mainstream RAG methods in answer accuracy (EM), F1 score, and multi-task comprehensive metrics, showcasing superior capabilities in complex knowledge integration and reasoning.

Overall, the main contributions of this paper are as follows:

(1) We propose the novel concept of Semantic Gist and the cognitive memory-inspired RAG framework CogitoRAG, which infers unstructured corpora into gist memory corpora and further encodes them into a multi-dimensional knowledge graph integrating entities relational facts and memory nodes.

(2) We design three core retrieval components: the Query Decomposition Module for simulating human cognitive decomposition of complex information, the Entity Diffusion Module for emulating episodic memory and importance judgment via topological relevance and entity frequency-based semantic diffusion, and the CogniRank algorithm for global context-aware reranking through weighted fusion of semantic and graph topological importance.

(3) Extensive experiments on six benchmarks demonstrate that CogitoRAG significantly outperforms state-of-the-art methods in QA, multi-hop reasoning, and multi-task generation tasks.

\section{Related work}
\subsection{Retrieval-Augmented Generation}
Retrieval-Augmented Generation (RAG) improves LLMs by grounding them in external knowledge, which helps reduce hallucinations and improve factual accuracy~\citep{lewis2020retrieval, jiang2023active}. However, the standard RAG paradigm, which decomposes documents into independent chunks, causing the retrieval process to fall into a local optima trap where it retrieves semantically related but contextually incomplete fragments~\citep{barnett2024seven}. While advanced frameworks have sought to address this, they often do so through sequential, multi-step processes. For instance, iterative architectures like ITER-RETGEN~\citep{shao2023iterative} were developed for multi-hop reasoning by progressively refining queries. Similarly, methods like Self-RAG~\citep{asai2024selfrag} introduce self-reflection and critique loops to validate retrieved context. Although these methods enhance system performance to a certain extent, they essentially treat reasoning as a linear chain reaction, focusing on local logical connections. Such retrieval paradigms, lacking macro-level comprehension, prevent models from forming deep cognition when faced with highly complex knowledge integration tasks. Therefore, the proposed CogitoRAG aims to reconstruct the RAG retrieval paradigm from a cognitive perspective. Its core lies in simulating human understanding and memory mechanisms: before retrieval occurs, a global gist memory base is first constructed through reasoning, ensuring that subsequent retrieval performs precise extraction guided by a complete semantic context.

\subsection{Graph Retrieval-Augmented Generation}

To address the limitations of processing linear text for complex tasks, recent research has shifted toward incorporating structured graph knowledge. Early methods, such as KAPING~\citep{baek2023kaping}, attempt to retrieve Knowledge Graph (KG) subgraphs and flatten them into text, which inevitably leads to the loss of relational context. Subsequently, approaches like ToG~\citep{sun2023think} and ToG2~\citep{ma2024think} empower LLM agents to perform symbolic reasoning via graph traversal; however, such agent-based traversal remains essentially a form of ``localized, step-by-step reasoning'' within the graph. While the evolution from linear Chain-of-Thought (CoT) to Graph-of-Thought (GoT) allows for synthesizing information across multiple paths~\citep{besta2025demystifying}, these methods remain constrained by the locality of path searching and lack a macroscopic grasp of the global semantic substrate. Brain-inspired models, such as HippoRAG~\citep{wang2024hipporag} and its successor HippoRAG2~\citep{wang2025hipporag2}, simulate the human hippocampal memory system by converting corpora into KGs and employing associative retrieval algorithms. Nevertheless, these methods rely heavily on explicit topological associations between entities. When encountering complex semantics, purely structural associations struggle to resolve entity ambiguity and contextual metaphors, often leaving retrieved triples in an isolated state that lacks the semantic coherence required to support complex reasoning. Consequently, CogitoRAG proposes a ``comprehension-before-memorization'' paradigm. This approach first extracts the Semantic Gist to capture both explicit and implicit semantic threads within the text, utilizing this as a foundation to construct a multi-dimensional knowledge graph. By leveraging a Entity Diffusion Module and the CogniRank algorithm, CogitoRAG achieves global semantic memory integration and associative recall within a single, unified process.

\begin{figure*}[t]
  \centering
  \vspace{-3.2cm} 
  \includegraphics[width=\textwidth]{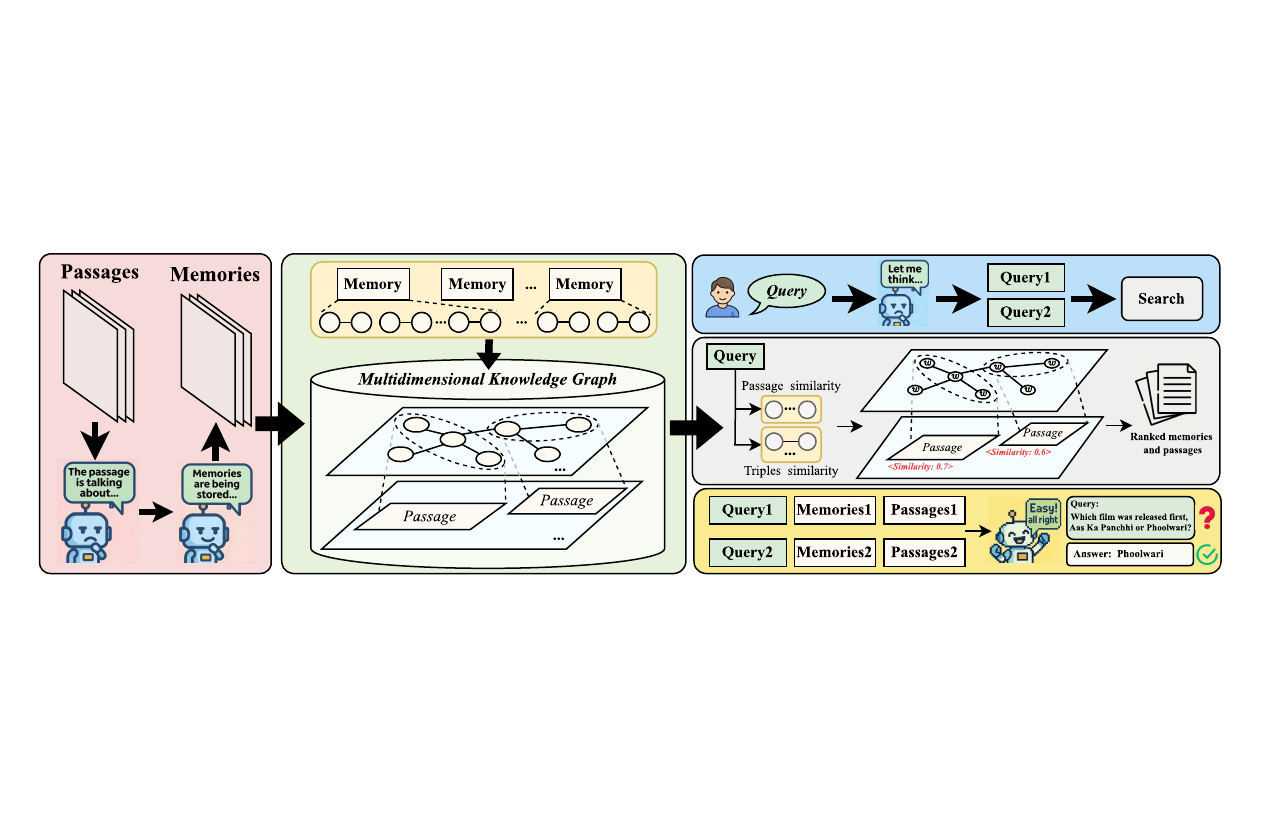}
  \vspace{-4cm} 
  \caption{The framework follows the "Understand Then Memory" paradigm: (1) Offline Indexing, it distills explicit facts and implicit relational logic from raw passages to construct a multi-dimensional knowledge graph that integrates semantic gist memories and contextual passages. (2) Online Retrieval, for a given query, it decomposes complex reasoning into sub-queries, then performs a unified cognitive diffusion process within the multi-dimensional knowledge graph. This enables global context-aware reranking of memories and passages via the CogniRank algorithm, delivering high-density evidential support to the generator for accurate and interpretable reasoning.}
  \label{fig:CogitoRAG}
\end{figure*}

\section{Preliminaries}

\paragraph{Retrieval-Augmented Generation} 
Retrieval-Augmented Generation~\citep{lewis2020retrieval,guu2020retrieval} is a widely adopted framework that integrates retrieval with generation to enhance the factuality and grounding of LLMs. Given a user query $q$, a retriever module $R$ identifies $k$ relevant documents $D_q = \{d_i\}_{i=1}^k$ from a large corpus $\mathcal{D} = \{d_i\}_{i=1}^N$, based on vector similarity. Formally:
\begin{equation}
D_q = \operatorname*{arg\,top\text{-}k} \left\{ \frac{E_d(d_i)^\top \cdot E_q(q)}{\|E_d(d_i)\|\|E_q(q)\|} \mid d_i \in \mathcal{D} \right\}
\end{equation}
where $E_d(\cdot)$ and $E_q(\cdot)$ denote the embedding functions for documents and the query, respectively.

The retrieved documents are concatenated with the query and passed to the LLM reader $\pi_\theta$ to generate a response:
\begin{equation}
y \sim \pi_\theta(y \mid q, D_q)
\end{equation}

\paragraph{Instruction-Following in RAG.} 
Modern RAG systems often operate under instruction-following settings~\citep{sun2023think,sun2024chain,wang2024hipporag,wang2025hipporag2}, where users provide explicit task directives $I = \{I_j\}_{j=1}^M$ along with the query $q$. The LLM $\pi_\theta$ is expected to generate output $y$ that not only answers the query using evidence from $D_q$, but also satisfies the constraints described in $I$:
\begin{equation}
y \sim \pi_\theta(y \mid q, D_q, I)
\end{equation}

\paragraph{Limitations of Standard RAG} 
Despite its effectiveness, standard RAG pipelines suffer from limitations in long-term factual retention, semantic coherence, and disambiguation due to noisy or redundant retrieval. These limitations motivate the need for structured and cognitively inspired memory augmentation techniques, such as CogitoRAG, introduced in the next section.

\section{Method}
\label{Method} 

\subsection{Overview}
As shown in Figure~\ref{fig:CogitoRAG}, CogitoRAG is a retrieval-augmented generation framework that simulates human-like memory organization and recall.
It addresses the lack of semantic integrity in conventional RAG systems by transforming raw passages into consolidated, disambiguated \emph{memories} and organizing them in a multi-dimensional knowledge graph for global, associative retrieval.
This design is inspired by cognitive theories of how humans encode and recall information: extracting a compact meaning representation from text, linking it to an episodic context, and applying an importance judgment mechanism during recall~\citep{reyna1995fuzzy,tulving1972episodic,kintsch1978toward}.

Formally, given a document collection $D=\{d_i\}$ and a user query $q$, CogitoRAG proceeds in two stages:
(1) \emph{offline indexing}, which parallels memory consolidation by segmenting documents into passages and extracting passage-grounded memories to build a graph integrating entities, facts, memories, and provenance passages; and
(2) \emph{online retrieval}, which mirrors recall by (i) optionally decomposing the query for coverage, (ii) diffusing activation over the passage--entity graph to obtain globally relevant evidence, and (iii) reranking passages via \textsc{CogniRank} and assembling passage--memory evidence pairs.
The final retrieved evidence is denoted as $\mathcal{E}(q)$, which is provided to the generator $\pi_\theta$ for answer synthesis:
\begin{equation}
y = \pi_\theta(q, \mathcal{E}(q)).
\end{equation}

\subsection{Offline Indexing}
The offline indexing stage simulates the organization and construction of human episodic memory, focusing on extracting passage-grounded cognitive memories that are consolidated, disambiguated, and KG-ready. It first splits unstructured documents into passage units, then reasons over each passage to capture explicit facts and implicit associations into consolidated memories. Finally, it constructs a multi-dimensional knowledge graph integrating entities, relational facts, and memory nodes based on these memories, providing a compositional and rich knowledge foundation for subsequent global and associative retrieval.

\subsubsection{Memory-Centric Knowledge Transformation}
Unlike conventional RAG systems that directly index raw text, CogitoRAG transforms unstructured passages into consolidated, disambiguated cognitive memories as a foundation for structured reasoning.

\paragraph{Graph Definition.} We build a multi-dimensional knowledge graph 
$\mathcal{G} = (\mathcal{V}, \mathcal{M}, \mathcal{E}, \mathcal{F}, \mathcal{P})$,
where $\mathcal{V}$ are entity nodes, $\mathcal{M}$ are memory nodes, $\mathcal{E}$ are relation types,
$\mathcal{F}$ are factual triples (facts), and $\mathcal{P}=\{p_j\}$ are passage nodes.

\paragraph{Document Segmentation.}
Each document $d_i \in D$ is segmented into semantically coherent passages $p_j$, forming $\mathcal{P}$.
Passages serve as the provenance-preserving evidence carrier for later answering, and also as the anchoring source from which memories and structured facts are derived.

\paragraph{Memory Extraction.}
For each passage $p$, we extract a KG-oriented memory $m(p)$ using an LLM with a prompt that outputs two fields: \texttt{<think>} and \texttt{<memory>}.
The \texttt{<think>} field documents the model's reasoning and cognitive processes regarding the text content, while the \texttt{<memory>} field retains the refined memory text that the model has comprehensively understood.
During graph construction, we only use the \texttt{<memory>} field and maintaining an explicit mapping between $m(p)$ and the original passage $p$ for provenance and reverse indexing.

Although the output format is unified, internally the passage processing can be roughly viewed as two coarse cases:
(1) When the passage is direct and factual with explicit relations, the extracted memory mainly performs light normalization, such as removing redundancy, resolving coreference, and rewriting implicit relations into explicit ones, which stabilizes downstream KG extraction.
(2) When deeper comprehension of the text is required, such as handling aliases or nicknames, implicit relations, narrative reasoning, or context-based disambiguation in complex medical knowledge, the extracted memory first performs strict text-based disambiguation. It then further extracts more critical implicit semantic content, and rewrites key entities, relations, and constraints into clearer and more extractable forms, without introducing any new facts beyond the original text.
In both cases, the \texttt{<memory>} content follows KG-friendly constraints, improving the extractability and reusability of nodes, edges, and triples.

\paragraph{Entity and Fact Extraction.}
We run an information extraction model $f_{\text{rel}}$ on each memory $m \in \mathcal{M}$ to identify entities and relational triples
$t=(v_h,r,v_t)$, where $v_h,v_t \in \mathcal{V}$ and $r\in\mathcal{E}$.
Each triple is aligned to its supporting memory node $m$, and can be traced back to the original passage via the $m(p)\leftrightarrow p$ mapping.
We also textualize each triple as a fact string
$f=\text{concat}(v_h,r,v_t),\; f\in\mathcal{F}$,
which serves as the similarity-based retrieval entry for selecting Top-$K$ facts in the online stage.

\subsubsection{Multi-dimensional Graph Construction.}
The final graph encodes three types of connections:
(1) structural links among entities induced by triples;
(2) memory nodes as intermediate semantic carriers connected to the entities and facts they contain;
and (3) passage nodes as provenance-preserving evidence bound to their corresponding memories for traceable answering.
As a result, the graph preserves not only structured relations, but also a reversible link between reusable semantic memories and the original textual evidence.

\paragraph{Vector Encoding.}
To support similarity retrieval and subsequent global reasoning, all objects are embedded into a shared vector space:
\begin{align}
e_v &= E_v(v), \quad v\in\mathcal{V}, \\
e_m &= E_m(m), \quad m\in\mathcal{M}, \\
e_r &= E_r(r), \quad r\in\mathcal{E}, \\
e_f &= E_f(\text{concat}(v_h,r,v_t)),\quad f\in\mathcal{F}, \\
e_p &= E_p(p), \quad p\in\mathcal{P}, \\
e_q &= E_q(q).
\end{align}
Here, $e_m$ represents the reusable semantic memory distilled from passages, and together with entities, facts, and provenance passages, forms the foundation for retrieval and reranking.
\subsection{Online Retrieval}
\label{sec:online_retrieval}

The online retrieval stage simulates the \emph{recall process}, where a query cue triggers associative activation over a structured memory network~\citep{kintsch1978toward}. 
CogitoRAG performs retrieval in three sequential steps: (1) \textbf{Query Decomposition Module} to improve coverage for multi-entity comparison questions; (2) an \textbf{Entity Diffusion Module} that initializes entity importance from top-ranked facts and propagates it over the entity graph (and further to passage nodes); and (3) \textbf{CogniRank} to rerank passages and assemble the final evidence for answer generation (original passages \emph{plus} their extracted memories).

\subsubsection{Query Decomposition Module}
Given a user query $q$, we first decide whether $q$ should be split into a small set of parallel sub-questions. 
This module targets queries that require retrieving information about two or more \emph{independent} entities. 
When splitting is triggered, an LLM produces up to $M$ short sub-questions $\mathcal{Q}(q)=\{q_1,\dots,q_m\}$; otherwise $\mathcal{Q}(q)=\{q\}$. 
In all cases, we maintain a mapping between the original query and the generated sub-questions to support traceability and later inspection.

\paragraph{Selecting the final top-$K$ passages.}
If no split is performed, we run the subsequent retrieval pipeline once on $q$ and return the top-$K$ passages.
If splitting is performed (typically $m=2$ in our setting), we run the pipeline separately for each $q_i$ and select the final top-$K$ passages using a \textbf{2+2+1} strategy for $K=5$:
we take the top-$2$ passages from $q_1$, the top-$2$ passages from $q_2$, and then add the single highest-scoring passage among the remaining candidates as a global complement.
This simple fusion ensures coverage across parallel sub-questions while preserving a strong global evidence item.

\subsubsection{Entity Diffusion Module}
For each query $q' \in \mathcal{Q}(q)$, we compute (i) fact-level similarity to obtain directly activated evidence, (ii) entity initial activation by combining fact evidence with an entity-frequency reward and a chunk-coverage penalty, and (iii) diffusion over the graph to incorporate global structural relevance. 
Importantly, we run the same diffusion mechanism to propagate activation not only among entity nodes but also onto \emph{passage} nodes, yielding a diffusion-derived passage relevance score for downstream reranking.

\paragraph{Fact similarity and top-$K$ facts.}
We encode the query as $e_{q'}=E_q(q')$ and compute cosine similarity against all textualized facts:
\begin{equation}
\sigma(q', f)=\frac{\langle e_{q'}, e_f\rangle}{\|e_{q'}\|\|e_f\|}, \quad f\in\mathcal{F}.
\end{equation}
We then select the top-$K$ facts:
\begin{equation}
\mathcal{F}_K(q')=\operatorname{TopK}_{f\in\mathcal{F}}\big(\sigma(q', f)\big).
\end{equation}

\paragraph{Entity evidence from facts.}
Instead of scoring an entity in isolation, we measure an entity's evidence strength through the top-$K$ facts in which it participates.
Let $\mathbf{1}[v\in f]$ indicate whether entity $v$ appears in fact $f$. We define:
\begin{equation}
\text{fact}(v)=
\frac{1}{|\{f \in \mathcal{F}_K(q') : v \in f\}|} 
\sum_{f \in \mathcal{F}_K(q')} \mathbf{1}[v \in f]\cdot \sigma(q', f),
\label{eq:fact_score}
\end{equation}
where $\text{fact}(v)$ aggregates query--fact similarities as entity-level evidence.

\paragraph{Entity-frequency reward (importance judgment).}
Motivated by the cognitive notion of \emph{importance judgment}~\citep{kintsch1978toward}, we further reward entities that are repeatedly supported by the top-$K$ facts.
Let $c_v$ be the hit count of entity $v$ in $\mathcal{F}_K(q')$. The reward factor is:
\begin{equation}
\text{reward}(v) = 1 + \alpha \big(1 - e^{-\beta c_v}\big),
\label{eq:reward}
\end{equation}
where $\alpha$ controls the reward magnitude and $\beta$ controls the saturation rate.

\paragraph{Chunk-coverage penalty and initial activation.}
Some entities can be linked to many passage chunks in the index (e.g., generic or highly frequent entities), which may inflate their importance.
To stabilize retrieval, we normalize by the number of chunks associated with the entity in the index.
Let $n_v$ be the number of chunks connected to entity $v$ (i.e., the size of the entity-to-chunk mapping); when $n_v>0$, we apply a penalty.
We define the initial activation (entity weight) multiplicatively:
\begin{equation}
\pi_0(v)=
\frac{\text{fact}(v)\cdot \text{reward}(v)}{\max(1, n_v)}.
\label{eq:pi0_mult}
\end{equation}
This definition couples evidence strength and frequency-based importance, while discouraging overly broad entities via chunk-coverage normalization.

\paragraph{Diffusion with restart on the passage--entity graph.}
Starting from $\pi_0$ on entity nodes, we perform spreading activation using a random-walk-with-restart update:
\begin{equation}
\pi_{t+1} = (1-\gamma)W^\top \pi_t + \gamma \pi_0,
\label{eq:diffusion}
\end{equation}
where $\gamma\in(0,1)$ is the restart probability and $W$ is the column-normalized adjacency matrix of the passage--entity graph (including both entity and passage nodes).
The process converges to a stationary distribution $\pi^\ast$, capturing both direct evidence (anchored by $\pi_0$) and global structural relevance (propagated through $W$).
We use the converged activation on each passage node as its diffusion-derived relevance score, denoted as $S_{\text{diff}}(p\mid q')$.
\subsubsection{CogniRank: Passage Reranking and Evidence Assembly}
\label{sec:cognirank}

Given the diffusion-derived passage score $S_{\text{diff}}(p\mid q')$, we further incorporate direct semantic matching between the (sub-)query and each passage.
Because $S_{\text{diff}}(p\mid q')$ and the embedding similarity $\sigma(q',p)$ can be on different numeric scales, we apply a simple min--max normalization over the candidate passage set $\mathcal{P}_c$ for each $q'$:
\begin{equation}
\operatorname{Norm}(x(p))=
\frac{x(p)-\min_{p'\in\mathcal{P}_c} x(p')}
{\max_{p'\in\mathcal{P}_c} x(p')-\min_{p'\in\mathcal{P}_c} x(p')+\delta},
\label{eq:minmax_norm}
\end{equation}
where $\delta$ is a small constant for numerical stability.

\paragraph{Single-parameter fusion for reranking.}
We fuse the two normalized signals with a single coefficient $\epsilon\in[0,1]$:
\begin{equation}
S(p\mid q') = \epsilon\cdot \operatorname{Norm}\!\left(S_{\text{diff}}(p\mid q')\right)
+(1-\epsilon)\cdot \operatorname{Norm}\!\left(\sigma(q',p)\right).
\label{eq:single_param_fusion}
\end{equation}
Here, $\epsilon=1$ reduces to a purely diffusion-based ranking (using only $S_{\text{diff}}$), while $\epsilon=0$ reduces to a purely similarity-based ranking (using only $\sigma$).
We rerank passages by $S(p\mid q')$ and select the top-$K$ passages for answering.

\paragraph{Evidence assembly with passage--memory pairing.}
During offline indexing, each passage $p$ is associated with a consolidated memory node $m(p)$ produced by the memory extraction module, and we store a one-to-one mapping $(p \leftrightarrow m(p))$.
In the answering stage, we assemble an evidence package by pairing each retrieved passage with its corresponding memory:
\begin{equation}
\mathcal{E}(q) = \left\{\big(p, m(p)\big)\ \middle|\ p \in \operatorname{TopK}_{p\in\mathcal{P}} S(p\mid q')\right\}.
\end{equation}
For each pair, the original passage provides verbatim grounding, while the memory provides a high-density representation with resolved references and clarified relations, facilitating more reliable answer synthesis.
The final evidence package $\mathcal{E}(q)$ is then formatted into the QA prompt for the generator to produce the answer.

\section{Experiment}
\label{Experiment}

\subsection{Experimental Setup}

\subsubsection{Baseline}
We compare CogitoRAG with several established retrieval methods and structure-enhanced RAG systems, including:  
(1) None, a QA model without external retrieval, relying solely on parametric knowledge;  
(2) NV-Embed-v2 (7B)~\citep{lee2024nv}, a strong dense retrieval baseline;  
(3) GraphRAG~\citep{graphrag2024}, a RAG framework augmented with graph-based corpus structuring;  
(4) LightRAG~\citep{guo2024lightrag}, a lightweight dual-level RAG approach; 
(5) RAPTOR~\citep{sarthi2024raptor}, a RAG method that builds a recursive tree structure from documents for hierarchical retrieval; 
(6) HippoRAG~\citep{wang2024hipporag}, a brain-inspired KG-based RAG framework with associative retrieval;  
(7) HippoRAG2~\citep{wang2025hipporag2}, an advanced successor that integrates dense--sparse retrieval with recognition memory; and  
\add{(8) ComoRAG~\citep{wang2025comorag}, a cognitive-inspired memory-organized RAG framework designed for stateful long narrative reasoning. }  
\add{(9) ToG2~\citep{ma2024think}, an agentic graph-traversal baseline that performs step-by-step reasoning on knowledge graphs.  }
All models employ the same QA reader (\texttt{gpt-4o-mini}), and CogitoRAG uses the same retriever (NV-Embed-v2) to ensure a fair comparison. We evaluate performance using three metrics: Exact Match (EM), F1 score, and Retrieval Time.

\subsubsection{Datasets}
To comprehensively evaluate the QA capability and retrieval performance of CogitoRAG, we categorize our benchmark datasets into Simple QA, which focuses on retrieving factual knowledge from a single passage using datasets like Natural Questions (NQ)~\citep{wang2024rear} and PopQA~\citep{mallen2022not}, and Multi-hop QA, which evaluates reasoning ability across multiple passages using datasets like MuSiQue~\citep{trivedi2022musique}, 2WikiMultihopQA~\citep{ho2020constructing}, and HotpotQA~\citep{yang2018hotpotqa}. 
We further include GraphBench~\citep{xiang2025use} to evaluate broader multi-task generation ability beyond QA (Novel/Medical; Fact Retrieval, Complex Reasoning, Contextual Summarization, and Creative Generation). \add{For GraphBench, we apply a simple document chunking step to build the index.}

Following the setup in HippoRAG~\citep{wang2024hipporag}, we uniformly sample 1,000 queries from each benchmark dataset, including PopQA~\citep{mallen2022not}, Natural Questions (NQ)~\citep{wang2024rear}, MuSiQue~\citep{trivedi2022musique}, 2WikiMultihopQA~\citep{ho2020constructing}, and HotpotQA~\citep{yang2018hotpotqa}.  
The detailed statistics of these datasets, including the number of queries and passages, are summarized in Table~\ref{tab:dataset-statistics}.

\begin{table*}[!t]
\centering
\caption{Dataset statistics}
\label{tab:dataset-statistics}
\begin{tabular}{lccccccc}
\toprule
Dataset & NQ & PopQA & MuSiQue & 2Wiki & HotpotQA & Novel & Medical \\
\midrule
Number of queries  & 1,000 & 1,000 & 1,000 & 1,000 & 1,000 & 2,010 & 2,062 \\
Number of passages & 9,633 & 8,676 & 11,656 & 6,119 & 9,811 & 1,108 & 224 \\
\bottomrule
\end{tabular}
\end{table*}

\paragraph{GraphBench.}
GraphBench~\citep{xiang2025use} is used for multi-task evaluation beyond QA; full benchmark results (Novel/Medical) are reported in Appendix~\ref{app:graphbench}.

\begin{table*}[!t]
\centering
\caption{EM and F1 for each retrieval method on five QA benchmarks, and GraphBench (Novel/Medical) multi-task \textbf{ACC} summary (FR/CR/CS)~\citep{xiang2025use}. Full GraphBench protocol metrics are reported in Appendix. Bold indicates the highest results. \textsuperscript{$\dagger$} indicates results are reproduced using the official code on our test datasets.}
\label{tab:overall}
\small
\setlength{\tabcolsep}{3.0pt}
\begin{tabular*}{\textwidth}{@{\extracolsep{\fill}}l *{16}{c}}
\toprule
\multirow{2}{*}{Method} 
& \multicolumn{2}{c}{NQ} 
& \multicolumn{2}{c}{PopQA} 
& \multicolumn{2}{c}{MuSiQue} 
& \multicolumn{2}{c}{2Wiki} 
& \multicolumn{2}{c}{HotpotQA}
& \multicolumn{3}{c}{Novel}
& \multicolumn{3}{c}{Medical} \\
\cmidrule(lr){2-3} \cmidrule(lr){4-5} \cmidrule(lr){6-7} \cmidrule(lr){8-9} \cmidrule(lr){10-11}
\cmidrule(lr){12-14} \cmidrule(lr){15-17}
& EM & F1 & EM & F1 & EM & F1 & EM & F1 & EM & F1
& FR & CR & CS & FR & CR & CS \\
\midrule
None                    & 35.20 & 52.70 & 16.10 & 22.70 & 11.20 & 22.00 & 30.20 & 36.30 & 28.60 & 41.00 & / & / & / & / & / & / \\
NV-Embed-v2~\citep{lee2024nv}          
                        & 43.50 & 59.90 & 41.70 & 55.80 & 32.80 & 46.00 & 54.40 & 60.80 & 57.30 & 71.00 & / & / & / & / & / & / \\
GraphRAG~\citep{graphrag2024}             
                        & 38.00 & 55.50 & 30.70 & 51.30 & 27.00 & 42.00 & 45.70 & 61.00 & 51.40 & 67.60 & 50.23 & 41.37 & 50.68 & 58.14 & 55.42 & 61.09 \\
LightRAG~\citep{guo2024lightrag}             
                        &  2.80 & 15.40 &  1.90 & 14.80 &  2.00 &  9.30 &  2.50 & 12.10 &  9.90 & 20.20 & 58.62 & 49.07 & 48.85 & 63.32 & 61.32 & 63.14 \\
RAPTOR~\citep{sarthi2024raptor}            
                        & 37.80 & 54.50 & 41.90 & 55.10 & 27.70 & 39.20 & 39.70 & 48.40 & 50.60 & 64.70 & 49.25 & 38.59 & 47.10 & 54.07 & 53.20 & 58.73 \\
HippoRAG~\citep{wang2024hipporag}     
                        & 37.20 & 52.20 & 42.50 & 56.20 & 24.00 & 35.90 & 59.40 & 67.30 & 46.30 & 60.00 & 52.93 & 38.52 & 48.70 & 56.14 & 55.87 & 59.86 \\
ToG2~\citep{ma2024think}\textsuperscript{$\dagger$}                 
                        & 36.60 & 43.70 & 20.30 & 22.60 & 10.30 & 17.50 & 31.90 & 36.40 & 31.50 & 41.20 & 42.15 & 37.20 & 43.86 & 55.11 & 40.24 & 64.88 \\
ComoRAG~\citep{wang2025comorag}\textsuperscript{$\dagger$}           
                        & 38.50 & 52.60 & 45.80 & \textbf{61.00} & 24.50 & 35.80 & 48.40 & 57.40 & 39.90 & 51.20 & 44.91 & 38.74 & 57.29 & 58.92 & 63.65 & 67.62 \\
HippoRAG2~\citep{wang2025hipporag2}       
                        & 43.40 & 60.00 & 41.70 & 55.70 & 35.00 & 49.30 & 60.50 & 69.70 & 56.30 & 71.10 & 60.14 & 53.38 & 64.10 & 66.28 & 61.98 & 63.08 \\
\rowcolor{blue!10}
\textbf{CogitoRAG (Ours)}                
                        & \textbf{51.55} & \textbf{63.76} & \textbf{50.94} & 59.94 & \textbf{43.20} & \textbf{53.95} & \textbf{69.90} & \textbf{76.20} & \textbf{61.62} & \textbf{75.00} & \textbf{62.44} & \textbf{56.27} & \textbf{70.99} & \textbf{72.57} & \textbf{74.40} & \textbf{75.51} \\
\bottomrule
\end{tabular*}
\end{table*}
\subsection{Results}

\subsubsection{Overall QA Performance}
We conduct a comprehensive comparison between \textsc{CogitoRAG} and all baseline methods on five benchmark datasets to evaluate overall QA performance.
The results are summarized in Table~\ref{tab:overall}, covering accuracy metrics (EM and F1), as well as GraphBench multi-task \textbf{ACC} (FR/CR/CS) under the Novel and Medical domains.
All baselines follow the same experimental setup described above.

As shown in Table~\ref{tab:overall}, \textsc{CogitoRAG} achieves the best EM on all five QA benchmarks, demonstrating strong and consistent gains across diverse QA settings.
The improvements are especially pronounced on multi-hop reasoning datasets.
For instance, on MuSiQue, \textsc{CogitoRAG} reaches 43.20 EM, outperforming the strong HippoRAG2 baseline by +8.20 EM.
This advantage is also clear on 2Wiki, where \textsc{CogitoRAG} achieves 69.90 EM, improving over HippoRAG2 by +9.40 EM.
On open-domain QA benchmarks such as NQ and PopQA, \textsc{CogitoRAG} also delivers substantial gains, indicating robust effectiveness across both single-hop and multi-hop settings.

Beyond QA, \textsc{CogitoRAG} also achieves the strongest GraphBench multi-task ACC summary across both domains.
On Novel, it obtains the best ACC on Fact Retrieval, Complex Reasoning, and Contextual Summarization.
On Medical, it further leads all compared methods on the same three task categories, showing strong generalization to graph-grounded generation settings beyond standard QA.
Full protocol metrics are reported in Appendix Table~\ref{tab:graphbench-novel} and Table~\ref{tab:graphbench-medical}.

These results provide direct empirical evidence that our framework effectively mitigates the localized reasoning limitations of conventional RAG systems.
By organizing corpus knowledge around \emph{understanding memory} and performing diffusion-based global reranking over a multi-dimensional graph, \textsc{CogitoRAG} supports more coherent evidence aggregation and stronger cross-context reasoning, which in turn yields more robust performance across both QA and long-form generation tasks.
\subsubsection{Graph Statistics: Node Count and Knowledge Density}
\label{sec:graph-statistics}

To provide a clearer view of graph complexity under different construction strategies, we report the \textbf{node count} across datasets in Table~\ref{tab:graph-nodes-compare}. 
We compare three graph variants: \textbf{Traditional KG}, \textbf{Summary}, and \textbf{Ours}.

\begin{table}[!t]
\centering
\caption{Node count comparison across three graph construction variants.}
\label{tab:graph-nodes-compare}
\setlength{\tabcolsep}{6pt}
\small
\begin{tabular}{lrrr}
\toprule
Dataset & Traditional KG & Summary & Ours \\
\midrule
PopQA      & 94,420  & 67,123  & 105,426 \\
NQ         & 96,537  & 84,430  & 116,876 \\
2Wiki      & 55,663  & 50,311  & 60,731 \\
MuSiQue    & 113,297 & 101,273 & 129,056 \\
HotpotQA   & 104,916 & 91,720  & 117,788 \\
Novel      & 20,250  & 13,019  & 40,541 \\
Medical    & 8,048   & 2,199   & 8,412 \\
\bottomrule
\end{tabular}

\vspace{3pt}
\footnotesize\textit{Note:} Traditional KG node counts are taken from HippoRAG2-constructed graphs on the five QA benchmarks (PopQA/NQ/2Wiki/MuSiQue/HotpotQA), and from GraphRAG-constructed graphs on Novel/Medical.
\end{table}

\noindent\textbf{Discussion.}
A larger node count does not necessarily imply a better graph.
What matters for downstream diffusion and reranking is the \emph{knowledge density}---i.e., how much answer-relevant semantics each node reliably encodes.

On fact-centric QA benchmarks including PopQA, NQ, 2Wiki, MuSiQue and HotpotQA, our construction can yield more nodes, because our \emph{understanding memory} explicitly captures both \textbf{explicit facts} and \textbf{implicit semantic cues} needed for reasoning (e.g., resolved references and implicit relations).
This makes more answer-critical units available as graph nodes, providing richer global signals for diffusion and reranking.

In contrast, on long-form or specialized corpora (Novel/Medical), summary-based graphs may inflate node counts by producing generic or fragmented representations, while our understanding memory aims to be more selective and structure-aware, leading to a smaller but more informative graph.
Overall, node count varies with data type, and our goal is not to maximize entities but to encode high-density semantic units that support global propagation and reranking, aligning with the improved QA accuracy observed in our ablations.
\subsubsection{Sensitivity Analysis: Hyperparameter Robustness}
\label{sec:sensitivity}

\paragraph{Entity-frequency reward $(\alpha,\beta)$.}
We study the sensitivity of the entity-frequency reward in Eq.~\eqref{eq:reward} on 2WikiMultiHopQA using two controlled sweeps:
(i) fixing $\alpha=2$ and scanning $\beta$ (saturation rate), and
(ii) fixing $\beta=1$ and scanning $\alpha$ (reward strength).
We report retrieval quality (Recall@K) and downstream QA performance (EM/F1).

\begin{table}[!t]
\centering
\caption{Fix $\alpha=2$ and scan $\beta$ on 2WikiMultiHopQA (all in \%).}
\label{tab:hparam-beta-scan-main}
\small
\setlength{\tabcolsep}{4.2pt}
\begin{tabular}{c|cc|ccc}
\toprule
$\beta$ & EM & F1 & R@10 & R@50 & R@200 \\
\midrule
0.5 & 68.10 & 74.66 & 94.50 & 96.00 & 96.65 \\
\textbf{1.0} & \textbf{69.90} & \textbf{76.20} & 94.55 & 96.00 & 96.65 \\
1.5 & 67.70 & 74.40 & 94.55 & 96.00 & 96.65 \\
2.0 & 67.80 & 74.50 & 94.53 & 96.03 & 96.67 \\
\bottomrule
\end{tabular}
\end{table}

\begin{table}[!t]
\centering
\caption{Fix $\beta=1$ and scan $\alpha$ on 2WikiMultiHopQA (all in \%).}
\label{tab:hparam-alpha-scan-main}
\small
\setlength{\tabcolsep}{4.0pt}
\begin{tabular}{c|cc|cccc}
\toprule
$\alpha$ & EM & F1 & R@5 & R@10 & R@50 & R@200 \\
\midrule
0.5 & 67.80 & 74.50 & 92.47 & 94.53 & 96.03 & 96.67 \\
1.0 & 68.00 & 74.56 & 92.40 & 94.55 & 96.00 & 96.65 \\
1.5 & 68.10 & 74.62 & 92.40 & 94.55 & 96.00 & 96.65 \\
\textbf{2.0} & \textbf{69.90} & \textbf{76.20} & \textbf{93.75} & \textbf{95.75} & \textbf{96.65} & \textbf{97.15} \\
2.5 & 68.20 & 74.75 & 92.40 & 94.55 & 96.00 & 96.65 \\

\bottomrule
\end{tabular}
\end{table}

Overall, performance varies only marginally across a wide range of $(\alpha,\beta)$.
Both retrieval (e.g., Recall@10/50/200) and end QA metrics (EM/F1) remain highly stable, indicating that the frequency-based reward introduces limited perturbation to the overall pipeline.
In our sweeps, $\beta=1.0$ and $\alpha\ge 2.0$ achieve the best EM/F1, suggesting that encouraging frequency-based importance is beneficial, while the system is not overly sensitive to the exact hyperparameter choice.

\paragraph{Sensitivity to fusion coefficient $\epsilon$.}
We study the sensitivity of the single-parameter fusion coefficient $\epsilon$ used in \textsc{CogniRank} (Eq.~\eqref{eq:single_param_fusion}) on 2WikiMultiHopQA.
This coefficient controls the trade-off between the diffusion-derived passage relevance score and direct passage--query semantic similarity after normalization.
In particular, $\epsilon=1$ corresponds to using \emph{only} the diffusion-derived score (pure diffusion-based ranking), while smaller $\epsilon$ increases the contribution of direct semantic matching.
Table~\ref{tab:hparam-epsilon} reports EM/F1 and Recall@K under four settings.

\begin{table}[!t]
\centering
\caption{Sensitivity analysis of fusion coefficient $\epsilon$ on 2WikiMultiHopQA (all in \%).}
\label{tab:hparam-epsilon}
\setlength{\tabcolsep}{4.5pt}
\small
\begin{tabular}{c|cc|ccccc}
\toprule
$\epsilon$ & EM & F1 & R@5 & R@10 & R@50 & R@100 & R@200 \\
\midrule
1.00 & 68.20 & 74.75 & 92.40 & 94.55 & 96.00 & 96.25 & 96.65 \\
\textbf{0.95} & \textbf{69.90} & \textbf{76.20} & \textbf{94.05} & \textbf{96.05} & \textbf{97.05} & \textbf{97.15} & \textbf{97.35} \\
0.90 & 69.80 & 75.79 & 93.95 & 95.85 & 97.00 & 97.05 & 97.25 \\
0.85 & 69.00 & 75.25 & 93.90 & 95.90 & 97.00 & 97.10 & 97.30 \\
\bottomrule
\end{tabular}
\end{table}

Overall, the results indicate that the fusion is relatively robust within a reasonable range of $\epsilon$.
Using diffusion-only ranking ($\epsilon=1.00$) is already strong, while introducing a small amount of semantic matching ($\epsilon\in[0.85,0.95]$) further improves both EM/F1 and Recall@K.
This suggests that diffusion-derived structural relevance should dominate, but lightweight semantic alignment remains beneficial for stabilizing the final evidence ordering.

\subsubsection{Ablation Study: Graph Construction and Retrieval Components}

To thoroughly evaluate the importance of each major design choice, we conduct ablation studies on the \textbf{MuSiQue} dataset from two perspectives:
\textbf{(graph construction} (how we preprocess passages before building the graph), and
\textbf{online retrieval} (which retrieval-time modules contribute to the final performance).
Unless otherwise specified, all ablations follow the same reader and retrieval setup as in Table~\ref{tab:overall}.

\paragraph{Ablations on graph construction (MuSiQue).}
We analyze how different graph construction strategies affect both downstream QA accuracy and graph complexity on MuSiQue.
We compare three variants:
(1) \textbf{Traditional KAG system}: directly construct the graph from full raw passages (no pre-compression);
(2) \textbf{Summary}: extract a short summary for each passage before graph construction;
(3) \textbf{Ours}: build the graph using our cognition-inspired understanding memory (semantic gist/memory) as described in Section~\ref{Method}.
We report \textbf{EM/F1} as well as the resulting graph scale (\#entity nodes and \#edges).

\begin{table}[!t]
\centering
\caption{Graph construction ablation on MuSiQue: QA performance and graph scale.}
\label{tab:ablation-graph-musique}
\setlength{\tabcolsep}{6pt}
\small
\begin{tabular}{lcccc}
\toprule
Method & EM(\%) & F1(\%) & \#Nodes & \#Edges \\
\midrule
Traditional KAG system & 35.0 & 49.3 & 113,297 & 1,304,605 \\
Summary                & 36.5 & 50.2 & 101,273 & 1,323,316 \\
\textbf{Ours}          & \textbf{43.2} & \textbf{53.95} & 129,056 & 1,784,432 \\
\bottomrule
\end{tabular}
\end{table}

\noindent
Overall, semantic preprocessing before graph construction improves QA performance (Summary vs.\ Traditional), and our understanding memory construction yields the best EM/F1 on MuSiQue.

\begin{table}[!t]
\centering
\caption{Retrieval-stage ablations on MuSiQue (EM/F1, in \%).}
\label{tab:ablation-retrieval}
\setlength{\tabcolsep}{8pt}
\small
\begin{tabular}{lcc}
\toprule
Method & EM(\%) & F1(\%) \\
\midrule
\textbf{Full system} & \textbf{43.2} & \textbf{53.95} \\
w/o EDF              & 35.0 & 49.30   \\
w/o CogniRank        & 36.5 & 50.23  \\
w/o QDM              & 41.7 & 53.01 \\
\bottomrule
\end{tabular}
\end{table}

\paragraph{Ablations on online retrieval components (MuSiQue).}
Next, we ablate key modules in the online retrieval stage on MuSiQue to quantify their individual contributions:
(1) \textbf{w/o EDF}: remove the Entity Diffusion Module (i.e., no global propagation over the entity graph);
(2) \textbf{w/o CogniRank}: remove the retrieval-time reranking module;
(3) \textbf{w/o QDM}: remove the \textit{Query Decomposition Module} and answer using a single-shot query without decomposition.

\subsection{Limitation and Future work}
\label{subsec:limitation-futurework}
Despite the significant advantages of CogitoRAG in complex knowledge integration and global reasoning tasks achieved through semantic gist extraction and comprehensive memory construction, several limitations remain to be addressed in future research.
First, the derivation of Gist Memory is inherently dependent on the reasoning capabilities of the underlying pre-trained language model. Its adaptability to cross-domain specialized terminology and ambiguous texts remains to be optimized. Specifically, the framework may encounter challenges in accurately extracting highly specialized implicit relational logic within specific domains (e.g., biomedicine or law). Future work will explore fine-tuning strategies on domain-specific datasets to enhance the precision and generalizability of gist extraction, ensuring a more robust capture of domain-specific semantic nuances.
Second, while our experiments demonstrate efficacy in controlled settings, the Entity Diffusion Module faces challenges in balancing reasoning efficiency and resource consumption when scaled to massive, real-world knowledge graphs (e.g., exceeding tens of millions of entities and relations). To improve deployment feasibility in large-scale practical scenarios, we plan to research lightweight and parallelizable diffusion algorithms. By incorporating strategies such as hierarchical indexing and dynamic topological pruning, we aim to significantly accelerate reasoning speeds while maintaining global semantic connectivity.
Finally, the dynamic iterative mechanism of the "Comprehension-Memorization" loop offers substantial room for further exploration. The current Gist Memory repository is constructed through an offline, static process, which lacks the capability to assimilate new domain knowledge in real-time or dynamically update existing semantic associations. Future research will focus on developing a dynamic cognitive memory architecture and designing incremental gist update algorithms. This will enable real-time expansion and self-optimization of the memory repository, ultimately fostering a closed-loop "comprehension-memorization-retrieval" adaptive mechanism in multi-turn interaction scenarios to more closely emulate human dynamic cognitive processes.

\section{Conclusion}
In this work, we propose CogitoRAG. The core of CogitoRAG lies in the extraction and evolution of Semantic Gist: during the offline indexing stage, the framework first distills unstructured corpora into gist memory corpora, which are then transformed into a multi-dimensional knowledge graph integrating entities, relational facts, and memory nodes. In the online retrieval stage, the Query Decomposition Module breaks complex queries into comprehensive sub-queries, mimicking the cognitive decomposition process humans use to handle complex information. Subsequently, the Entity Diffusion Module performs associative retrieval across the graph guided by structural relevance and an entity-frequency reward mechanism. Meanwhile, the proposed CogniRank algorithm precisely reranks candidate passages by fusing diffusion-derived scores with semantic similarity, and finally delivers high-density information support to the generator in a passage-memory pairing format. Experimental results on five mainstream QA benchmarks and the GraphBench multi-task generation task demonstrate that CogitoRAG significantly outperforms state-of-the-art RAG methods, exhibiting superior performance in complex knowledge integration and reasoning tasks. This work demonstrates the great potential of integrating cognitive principles into RAG design, providing a novel perspective for overcoming the local optimization trap in traditional retrieval.

\section{Ethics statement}
The authors have adhered to the ACM Code of Ethics and Professional Conduct. This research is based on publicly available datasets, and their use complies with the respective licenses and terms of service. The study did not involve human subjects, and no new data containing personally identifiable information was collected. The authors declare no competing interests or conflicts of interest. We are committed to responsible and transparent research practices.

\section{Reproducibility statement}
To ensure the reproducibility of our research, we have made our code, data, and experimental setup fully available. The complete source code for our proposed CogitoRAG framework, along with scripts to replicate all experiments, is provided as supplementary material and will be released on GitHub upon publication. A detailed description of our model architecture and the CogniRank algorithm is presented in Section ~\ref{Method}. The public datasets used in our evaluation are listed in Section ~\ref{Experiment}.

\bibliographystyle{ACM-Reference-Format}
\bibliography{sample-base}
\appendix

\newpage 
\section{Appendix}
\label{app_exp_details} 

\subsection{Evaluation Metrics}
\label{app:metrics}

We report evaluation metrics under two complementary settings: (i) standard QA benchmarks, where answers are short spans/phrases; and (ii) GraphBench multi-task generation benchmarks, where the model produces task-specific outputs evaluated by the benchmark protocol.

\paragraph{QA Benchmarks: Exact Match and F1.}
To evaluate the performance of our system on question answering datasets, we adopt two widely used metrics: Exact Match (EM) and F1 score. 
Exact Match (EM) measures the percentage of predictions that match any one of the ground-truth answers exactly.  
F1 Score measures the overlap between the predicted and reference answer at the token level, treating the problem as a bag-of-tokens comparison.  

Formally, for a predicted answer $p$ and a reference answer $r$:
\[
\text{Precision} = \frac{|p \cap r|}{|p|}, \quad
\text{Recall} = \frac{|p \cap r|}{|r|}
\]
\[
\text{EM} =
\begin{cases}
1, & \text{if } p = r \\
0, & \text{otherwise}
\end{cases}
\]
\[
\text{F1} = \frac{2 \cdot \text{Precision} \cdot \text{Recall}}{\text{Precision} + \text{Recall}}
\]
When multiple reference answers exist, the maximum EM or F1 across all references is taken.

\paragraph{GraphBench: Multi-task Generation Metrics.}
For GraphBench~\citep{xiang2025use}, we follow the official evaluation protocol and report metrics for each task category, including Fact Retrieval (FR), Complex Reasoning (CR), Contextual Summarization (CS), and Creative Generation (CG).
GraphBench uses multiple task-dependent metrics, including:
\textbf{ACC} (accuracy), \textbf{ROUGE-L} (lexical overlap based on longest common subsequence), \textbf{Coverage (Cov)} (coverage of required key points/attributes), and \textbf{FS} (the benchmark-reported score for Creative Generation under its protocol).
Since \textbf{ACC} is the primary indicator for overall task success and is consistently available across task categories, we report \textbf{ACC} as the main summary metric in the main paper (e.g., Table~\ref{tab:overall}). 
For completeness, we provide the full GraphBench results with all protocol metrics (ACC/ROUGE-L/Cov/FS) in Appendix Table~\ref{tab:graphbench-novel} and Table~\ref{tab:graphbench-medical}.

\subsection{Graph Composition Statistics}
\label{app:graph-composition}

To further characterize the constructed graphs, we report a detailed breakdown of graph components and extraction outcomes across all seven datasets in Table~\ref{tab:graph-composition-stats}. 
In addition to the overall graph size (\#nodes and \#edges), we include the corpus chunk statistics (\#chunks), the counts of key graph elements (\#entities and \#facts), and the OpenIE extraction totals (number of processed documents, extracted entities, and extracted triples). 
These statistics provide a concrete view of how much structured knowledge is induced from each dataset and support reproducibility of our graph construction pipeline.

\begin{table*}[!t]
\centering
\caption{Graph composition statistics across seven datasets. \#Chunks denotes the number of segmented passages/chunks used for graph construction. \#Entity and \#Fact denote the number of entity nodes and factual units (textualized triples) in the final graph. OpenIE totals report the extraction outcomes before graph consolidation.}
\label{tab:graph-composition-stats}
\small
\setlength{\tabcolsep}{4pt}
\begin{tabular*}{\textwidth}{@{\extracolsep{\fill}}lrrrrrrrr}
\toprule
Dataset & \#Nodes & \#Edges & \#Chunks & \#Entity & \#Fact & OpenIE Docs & OpenIE Entities & OpenIE Triples \\
\midrule
PopQA            & 105,426 & 1,374,946 & 8,676  & 96,750  & 124,601 & 8,676  & 94,847  & 129,912 \\
NQ              & 116,876 & 1,599,520 & 9,633  & 107,243 & 134,448 & 9,633  & 118,657 & 139,625 \\
Novel           & 40,541  & 336,188   & 1,108  & 39,433  & 37,256  & 1,108  & 18,705  & 37,310  \\
MuSiQue         & 129,056 & 1,784,432 & 11,656 & 117,400 & 141,998 & 11,656 & 122,674 & 144,472 \\
Medical         & 8,412   & 103,248   & 224    & 8,188   & 8,667   & 224    & 4,983   & 9,158   \\
HotpotQA        & 117,788 & 1,499,380 & 9,811  & 107,977 & 134,898 & 9,811  & 116,799 & 135,810 \\
2WikiMultiHopQA & 60,731  & 893,362   & 6,119  & 54,612  & 69,167  & 6,119  & 62,418  & 69,479  \\
\bottomrule
\end{tabular*}
\end{table*}

\subsection{GraphBench Multi-task Generation Evaluation (Novel / Medical)}

\label{app:graphbench}
To evaluate generalization beyond multi-hop QA, we adopt GraphBench~\citep{xiang2025use}, which covers two domains (Novel and Medical) and multiple task types of varying complexity, including Fact Retrieval, Complex Reasoning, Contextual Summarization, and Creative Generation.
Table~\ref{tab:graphbench-novel} and Table~\ref{tab:graphbench-medical} report the full results.
For a concise summary, we also include the average ACC across the four task types (Fact / Reasoning / Summarization / Creative) as the last column in each table.

\begin{table*}[!t]
\centering
\caption{GraphBench results on the Novel dataset. Metrics follow the benchmark protocol.}
\label{tab:graphbench-novel}
\small
\setlength{\tabcolsep}{4pt}
\begin{tabular*}{\textwidth}{@{\extracolsep{\fill}}lccccccccc}
\toprule
Model
& \multicolumn{2}{c}{Fact Retrieval}
& \multicolumn{2}{c}{Complex Reasoning}
& \multicolumn{2}{c}{Contextual Summarization}
& \multicolumn{2}{c}{Creative Generation}
& Avg \\
\cmidrule(lr){2-3}\cmidrule(lr){4-5}\cmidrule(lr){6-7}\cmidrule(lr){8-9}\cmidrule(lr){10-10}
& ACC & ROUGE-L & ACC & ROUGE-L & ACC & Cov & ACC & Cov & ACC \\
\midrule
RAG (w/o rerank) & 58.76 & 37.35 & 41.35 & 15.12 & 50.08 & 82.53 & 41.52 & 37.84 & 47.93 \\
RAG (w rerank)   & 60.92 & 36.08 & 42.93 & 15.39 & 51.30 & 83.64 & 38.26 & 40.04 & 48.35 \\
MS-GraphRAG      & 49.29 & 26.11 & 50.93 & 24.09 & 64.40 & 75.58 & 39.10 & 35.65 & 50.93 \\
HippoRAG         & 52.93 & 26.65 & 38.52 & 11.16 & 48.70 & 85.55 & 38.85 & 38.97 & 44.75 \\
HippoRAG2        & 60.14 & 31.35 & 53.38 & 33.42 & 64.10 & 70.84 & 48.28 & 30.95 & 56.48 \\
LightRAG         & 58.62 & 35.72 & 49.07 & 24.16 & 48.85 & 63.05 & 23.80 & 25.01 & 45.09 \\
Fast-GraphRAG    & 56.95 & 35.90 & 48.55 & 21.12 & 56.41 & 80.82 & 46.18 & 36.99 & 52.02 \\
RAPTOR           & 49.25 & 23.74 & 38.59 & 11.66 & 47.10 & 82.33 & 38.01 & 35.88 & 43.24 \\
Lazy-GraphRAG    & 51.65 & 36.97 & 49.22 & 23.48 & 58.29 & 76.94 & 43.23 & 39.74 & 50.60 \\
\textbf{CogitoRAG} & \textbf{62.44} & 32.88 & \textbf{56.27} & 30.25 & \textbf{70.99} & 41.88 & \textbf{59.56} & 36.42 & \textbf{62.32} \\
\bottomrule
\end{tabular*}
\end{table*}

\begin{table*}[!t]
\centering
\caption{GraphBench results on the Medical dataset. Metrics follow the benchmark protocol.}
\label{tab:graphbench-medical}
\small
\setlength{\tabcolsep}{4pt}
\begin{tabular*}{\textwidth}{@{\extracolsep{\fill}}lccccccccc}
\toprule
Model
& \multicolumn{2}{c}{Fact Retrieval}
& \multicolumn{2}{c}{Complex Reasoning}
& \multicolumn{2}{c}{Contextual Summarization}
& \multicolumn{2}{c}{Creative Generation}
& Avg \\
\cmidrule(lr){2-3}\cmidrule(lr){4-5}\cmidrule(lr){6-7}\cmidrule(lr){8-9}\cmidrule(lr){10-10}
& ACC & ROUGE-L & ACC & ROUGE-L & ACC & Cov & ACC & Cov & ACC \\
\midrule
RAG (w/o rerank) & 63.72 & 29.21 & 57.61 & 13.98 & 63.72 & 77.34 & 58.94 & 57.87 & 61.00 \\
RAG (w rerank)   & 64.73 & 30.75 & 58.64 & 15.57 & 65.75 & 78.54 & 60.61 & 58.72 & 62.43 \\
MS-GraphRAG      & 38.63 & 26.80 & 47.04 & 21.99 & 41.87 & 22.98 & 53.11 & 39.42 & 45.16 \\
HippoRAG         & 56.14 & 20.95 & 55.87 & 13.57 & 59.86 & 62.73 & 64.43 & 65.56 & 59.08 \\
HippoRAG2        & 66.28 & 36.69 & 61.98 & 36.97 & 63.08 & 46.13 & 68.05 & 51.54 & 64.85 \\
LightRAG         & 63.32 & 37.19 & 61.32 & 24.98 & 63.14 & 51.16 & 67.91 & 51.58 & 63.92 \\
Fast-GraphRAG    & 60.93 & 31.04 & 61.73 & 21.37 & 67.88 & 52.07 & 65.93 & 44.73 & 64.12 \\
RAPTOR           & 54.07 & 17.93 & 53.20 & 11.73 & 58.73 & 78.28 & 62.38 & 63.63 & 57.10 \\
Lazy-GraphRAG    & 60.25 & 31.66 & 47.82 & 22.68 & 57.28 & 55.92 & 62.22 & 43.79 & 56.89 \\
\textbf{CogitoRAG} & \textbf{72.57} & \textbf{54.55} & \textbf{74.40} & \textbf{38.56} & \textbf{75.51} & 51.33 & \textbf{68.07} & 42.50 & \textbf{72.64} \\
\bottomrule
\end{tabular*}
\end{table*}

Across both domains, CogitoRAG achieves the strongest performance on high-difficulty tasks (notably Complex Reasoning and Contextual Summarization),and also attains the highest average ACC, indicating superior robustness in broader multi-task scenarios beyond QA.

\subsection{Comparison: Passages vs.\ Triples vs.\ Passages+Memory as Answering Carrier}

To assess how different \emph{answering carriers} affect the final QA performance, we conduct a controlled comparison on the final answering stage.
We consider three settings:

(1) \textbf{Triples-as-Carrier}: the selected triples (facts) are directly provided to the QA reader as supporting evidence, without converting them back to raw text.

(2) \textbf{Passages-as-Carrier}: the selected triples are reverse-indexed to their source passages, and the \emph{original passage texts} are included in the QA prompt for answer generation.

(3) \textbf{Passages+Memory-as-Carrier}: in addition to the original passages, we further inject our extracted \emph{understanding memory} into the QA prompt.
Concretely, this memory corresponds to the semantic gist-derived cognitive representation constructed in our offline stage (Section~\ref{Method}), i.e., concise gist/memory statements distilled by the LLM that preserve the central proposition, key entities, and relational intent of the supporting context.
This additional memory serves as a compact semantic scaffold that helps the QA reader disambiguate nuanced modifiers, resolve implicit relations, and better align the retrieved evidence with the query intent.

\begin{table}[!t]
  \centering
  \caption{Comparison of different answering carriers on MuSiQue (EM/F1, in \%).}
  \label{tab:carrier-comparison}
  \setlength{\tabcolsep}{10pt}
  \small
  \begin{tabular}{lcc}
    \toprule
    Method & EM & F1 \\
    \midrule
    Triples-as-Carrier            & 14.90 & 23.06 \\
    Passages-as-Carrier           & 36.50 & 50.23 \\
    Passages+Memory-as-Carrier    & \textbf{43.20} & \textbf{53.95} \\
    \bottomrule
  \end{tabular}
\end{table}

Table~\ref{tab:carrier-comparison} shows that using \textbf{original passages} as the answering carrier substantially outperforms feeding \textbf{triples only},
indicating that lexical fidelity and local narrative context in raw text are crucial for precise span-style answering on MuSiQue.
More importantly, augmenting the passages with our \textbf{understanding memory} yields a further gain (+6.70 EM / +3.72 F1 over passages-only),
suggesting that the distilled semantic gist provides complementary high-level cues that help the QA reader capture the intended semantics beyond surface-form matching.

\subsection{Comparison with Structure-Augmented RAG Methods: Token Usage}
\label{app:token-usage}

We report the total number of input and output tokens consumed in the \emph{offline indexing} stage on the MuSiQue corpus (11,656 passages).
\textsc{CogitoRAG} is measured using 4omini, while numbers for other methods follow their reported experimental settings (e.g., Llama-based backbones) when available.
Note that token counts are backbone- and tokenizer-dependent; therefore, these statistics should be interpreted as a cost reference rather than a strictly controlled comparison.
Table~\ref{tab:app-token-usage} summarizes the results and shows relative proportions normalized by \textsc{CogitoRAG} (100\%).

\begin{table*}[!t]
\centering
\caption{Token usage of different structure-augmented RAG methods for indexing the MuSiQue corpus (11,656 passages) and their relative proportions (normalized by \textsc{CogitoRAG}).}
\label{tab:app-token-usage}
\setlength{\tabcolsep}{8pt}
\small
\begin{tabular}{lccccc}
\toprule
 & \textbf{\textsc{CogitoRAG}} & HippoRAG 2 & RAPTOR & LightRAG & GraphRAG \\
\midrule
Input Tokens  & \textbf{24.2M (100.0\%)} & 9.2M (38.0\%)  & 1.7M (7.0\%)   & 68.5M (283.0\%)  & 115.5M (477.1\%) \\
Output Tokens & \textbf{6.0M (100.0\%)}  & 3.0M (50.0\%)  & 0.2M (3.3\%)   & 18.3M (304.8\%)  & 36.1M (601.4\%) \\
\bottomrule
\end{tabular}
\end{table*}

\paragraph{Analysis.}
From the overall token cost, \textsc{CogitoRAG} is substantially more token-efficient than heavier structure augmented pipe-lines such as LightRAG and GraphRAG on MuSiQue.
Meanwhile, \textsc{CogitoRAG} consumes more indexing tokens than HippoRAG 2 and RAPTOR.
The primary reason is that \textsc{CogitoRAG} explicitly introduces a \emph{memory construction} stage in offline indexing: before graph building, raw passages are transformed into consolidated and disambiguated memories (e.g., resolving references and improving semantic integrity), which adds additional generation cost.
This deliberate trade-off yields cleaner and more stable semantic units, providing stronger signals for subsequent global diffusion retrieval and re-ranking.

\subsection{Gist and Multi-dimensional Knowledge Graph}
Through systematic analysis and experiments, this study reveals the importance and synergistic effects of each core component in the proposed cognition-inspired framework. The primary finding is that the inferential extraction of semantic gist serves as a crucial prerequisite for constructing a high-quality knowledge foundation. Ablation studies clearly demonstrate that bypassing this inferential step and directly building a knowledge graph based on the surface-level information of raw text will inevitably lead to semantic deviation during the retrieval phase. For instance, when responding to the query, ``Which film and television works starring actor Chris Evans include newcomers in their cast?'', a system lacking gist inference will likely fail. A vector-based RAG might retrieve irrelevant documents about ``drumming newcomers'' due to spurious semantic similarity , while a standard Knowledge Graph-indexed RAG might correctly link ``Chris Evans'' to the film ``The Newcomers'' but cannot verify if the cast members were actual newcomers, thus failing to answer the query's core intent. This retrieval bias, caused by ``gist deficiency,'' highlights the necessity of simulating human ``comprehending memory''. CogitoRAG, through its inferential step, grasps the true semantic gist—that ``newcomers'' refers to actors who were ``in the early stages of their acting careers''. More importantly, through carefully designed inferential steps, CogitoRAG distinguishes between key facts for verbatim retention (e.g., names like `Chris Evans', `Paul Dano' ) and semantic cores requiring refinement (the concept of a `newcomer actor' ), achieving a more human-like grasp of knowledge.

Based on this profound understanding of semantic gist, this study further verifies the effectiveness of constructing a Multidimensional Knowledge Graph  that integrates multi-dimensional information. This graph not only stores entity-relationship triples (e.g., `Chris Evans' - `stars in' - `The Newcomers' ) but, more importantly, anchors the semantic gist from which each triple is derived (e.g., `Paul Dano newcomer' ) and the corresponding raw passage. This design allows the system, when generating final answers, to leverage the graph's cross-document association capability while tracing back to the original context for accuracy. The case study clearly demonstrates that for complex questions, the model uses gist-guided retrieval to prioritize and integrate raw paragraphs rather than relying on discrete, context-poor triples. For example, to answer the query about Chris Evans' co-stars, the system retrieves the full contextual passage, ``Paul Dano who were in the early stages of their acting careers''. This practice of returning to the source text ensures the generated response is not only accurate but also contextually rich and fluent, effectively avoiding the unnatural generation and context deficiency that arise from relying solely on structured knowledge.

\begin{table*}[t]
\centering
\small
\renewcommand{\arraystretch}{1.25}
\begin{tabular}{p{0.97\linewidth}}
\noalign{\vskip 6pt}\hline\noalign{\vskip 6pt}
\textbf{Query:}
Which film has the director born later, \emph{Arrête Ton Cinéma} or
\emph{Agni (2004 Film)}? \\[6pt]
\noalign{\vskip 6pt}\hline\noalign{\vskip 6pt}
\textbf{Sub-questions:}
\begin{itemize}\setlength{\itemsep}{3pt}
\item What is the birth year of the director of \emph{Arrête Ton Cinéma}?
\item What is the birth year of the director of \emph{Agni (2004 Film)}?
\end{itemize} \\ 
\noalign{\vskip 6pt}\hline\noalign{\vskip 6pt}
\textbf{Retrieval for SubQ$_1$ — Arrête Ton Cinéma} \\[4pt]
\textbf{Retrieved Documents (Condensed):}
\begin{itemize}\setlength{\itemsep}{3pt}
\item \emph{Arrête ton cinéma}: directed by Diane Kurys.
\item \emph{Diane Kurys}: born 3 December 1948.
\end{itemize}
\textbf{Key Evidence:}
Director = Diane Kurys; Birth year = 1948. \\[6pt] \\
\textbf{Retrieval for SubQ$_2$ — Agni (2004 Film)} \\[4pt]
\textbf{Retrieved Documents (Condensed):}
\begin{itemize}\setlength{\itemsep}{3pt}
\item \emph{Agni (2004 film)}: directed by Swapan Saha.
\item \emph{Swapan Saha}: born 10 January 1930.
\end{itemize}
\textbf{Key Evidence:}
Director = Swapan Saha; Birth year = 1930. \\[6pt]
\noalign{\vskip 6pt}\hline\noalign{\vskip 6pt}
\textbf{Evidence Synthesis } \\[4pt]
Compare birth years: 1948 vs.\ 1930 $\Rightarrow$ 1948 is later. \\[4pt]
\textbf{Final Answer:}
\textbf{Arrête Ton Cinéma}. \\[6pt]
\noalign{\vskip 6pt}\hline\noalign{\vskip 6pt}
\end{tabular}
\caption{
Case Study on Comparative Reasoning with Split.
Our system decomposes the query into two sub-questions, retrieves focused
evidence for each entity, and synthesizes the final answer.
}
\label{tab:split_rag_case}
\end{table*}

\begin{table*}[t]
\centering
\scriptsize
\setlength{\tabcolsep}{3pt}
\renewcommand{\arraystretch}{1.15}

\newcolumntype{Y}{>{\raggedright\arraybackslash}X}

\begin{tabularx}{\textwidth}{l Y Y Y Y}
\hline
 & \textbf{Question} & \textbf{NV-Embed-v2 Results} & \textbf{CogitoRAG Filtered Triples} & \textbf{CogitoRAG Results} \\
\hline

\textbf{Simple QA}
& When did Lothair II's mother die?
&
1. Lothair II \newline
2. Teutberga \newline
3. Waldrada of Lotharingia
&
(\textbf{Ermengarde of Tours}, date of death, \textbf{20 March 851}) \newline
(Lothair II, mother, \textbf{Ermengarde of Tours})
&
1. \textbf{Ermengarde of Tours} \newline
2. \textbf{20 March 851}
\\
\hline

\textbf{Multi-Hop QA}
& What is the place of birth of the performer of song Changed It?
&
1. \textbf{Changed It} \newline
2. Nicki Minaj \newline
3. Did It On'em
&
(\textbf{Changed It}, performed by, \textbf{Nicki Minaj}) \newline
(\textbf{Nicki Minaj}, place of birth, \textbf{Saint James, Port of Spain}) \newline
(\textbf{Saint James}, located in, \textbf{Port of Spain})
&
1. \textbf{Nicki Minaj} \newline
2. \textbf{Saint James, Port of Spain} \newline
3. \textbf{Port of Spain}
\\
\hline
\end{tabularx}

\caption{Comparison of NV-Embed-v2 and CogitoRAG retrieval behavior. Bolded items denote titles of supporting passages.}
\end{table*}

\section{LLM Usage}
LLMs were used solely to enhance the linguistic quality of the manuscript, aiding in refining language, improving readability, ensuring clarity, rephrasing sentences, checking grammar, and boosting text flow, without any involvement in scientific content or data analysis. The authors fully take responsibility for the manuscript's content, and have ensured all content follows ethical guidelines and avoids plagiarism or scientific misconduct.

\section{Implementation Details}

\begin{mycode}{Python}{Prompt for NER}
"""Your task is to extract named entities from the given paragraph.
Respond with a JSON list of entities.
Note: Include specific factual details such as temporal information, geographical information, and numerical data alongside standard named entities."""
\end{mycode}
\begin{mycode}{Python}{Prompt for Triple}
"""Your task is to construct an RDF (Resource Description Framework) graph from the given passages and named entity lists.
Respond with a JSON list of triples, with each triple representing a relationship in the RDF graph.
Pay attention to the following requirements:
- Each triple should contain at least one, but preferably two, of the named entities in the list for each passage.
- Clearly resolve pronouns to their specific names to maintain clarity.
- Preserve specific factual details (temporal, geographical, numerical information) in the relationships when relevant."""
\end{mycode}

\begin{mycode}{Python}{Prompt for Memory}
"""You are a RAG memory-writing analyst. Your output will be used to build a knowledge graph (KG), so it must be faithful, extractable, and reusable.

In your internal thinking, you may roughly view the input as two kinds (ONLY for thinking; do NOT explicitly label the type in the final output):
- Direct factual text: information is clear and relations are explicit; usually only light cleaning is needed (remove redundancy, resolve coreference, make relations explicit).
- Understanding-required text: logic is more complex / contains references or nicknames / needs context-based disambiguation (e.g., metaphors, aliases, implied meaning, relation changes). You should first understand and make the key relations explicit, but must stay strictly grounded in the passage (no invention).
Task steps (the final output MUST include both parts):
Step 1: In <think>, write a brief ``memory strategy'': whether disambiguation/understanding is needed; which key entities/relations/constraints/time points to keep; how to handle references and redundancy.
Step 2: In <memory>, write the ``final memory content'': high information density and clear structure, easy for later triple/node-edge extraction (include entities, time, attributes/actions, explicit relations when possible).

You MUST output strictly in the following format, and output ONLY these two fields:
<think>
...
</think>
<memory>
...
</memory>

Rules:
1. Both fields must be present and non-empty.
2. Use only information supported by the passage. Do NOT invent or fill in missing dates, numbers, causes, or background.
3. <think> should not be a long chain-of-thought; only strategy bullet points or concise notes.
4. If the passage uses coarse time granularity, <memory> must keep the same granularity (do not over-specify).
5. CRITICAL for KG: Do NOT use pronouns (he, she, it, they, his, her, its, their, him, them) in <memory>. Always repeat the full entity name so each sentence is self-contained and unambiguous for triple extraction. This is essential for multi-hop reasoning.
"""
\end{mycode}
\begin{mycode}{Python}{Prompt for Query Decomposition Module}
"""You are a query decomposition assistant for document retrieval.
Your task:
- Decide whether the question should be split into simpler sub-questions.
- If splitting is necessary, set "split" to true and provide up to {max_splits} short sub-questions.
- If no split is needed, set "split" to false and leave "sub_questions" as an empty list.
Split the question ONLY when:
- answering requires retrieving information about two or more independent entities.
- the question involves comparison (e.g., earlier, later, same, different, both).
- the question can be naturally decomposed into parallel fact-finding queries.
Do NOT split when:
- the question asks about a single entity through a chain of relations.
- the question involves family or social relations.
- splitting would introduce ambiguous references (e.g., "the director").
Rules:
- Output ONLY valid JSON.
- Do NOT include explanations, markdown, or extra text.
- Use exactly this format:
{{"split": true|false, "sub_questions": ["..."]}}"""
\end{mycode}
\end{document}